\documentclass[a4paper]{llncs}
\usepackage{graphicx}
\usepackage{array,longtable}
\usepackage{algorithmic}
\usepackage{gensymb}
\usepackage{amsmath}
\usepackage{amsfonts}
\usepackage{subcaption}
\usepackage{mathtools}
\usepackage{cite}
\usepackage{xcolor}

\usepackage{multirow}
\usepackage[para,flushleft]{threeparttable}

\DeclarePairedDelimiter\floor{\lfloor}{\rfloor}

\usepackage{algorithm}

\begin{document}
\title{Learning Ensembles of Anomaly Detectors on Synthetic Data\thanks{The research was partially supported by the Russian Foundation for Basic Research grants 16-29-09649 ofi m.}}
%
%
\author{D. Smolyakov\inst{1} \and
N. Sviridenko\inst{1} \and V. Ishimtsev\inst{1} \and
E. Burikov\inst{2} \and E. Burnaev\inst{1}}
\authorrunning{D. Smolyakov et al.}
%
\institute{Skolkovo Institute of Science and Technology, Moscow Region, Russia \and
PO–AO “Minimaks-94”, Moscow, Russia\\
\email{Dmitrii.Smoliakov@skolkovotech.ru, Nadezda.Sviridenko@skolkovotech.ru, V.Ishimtsev@skoltech.ru,  Burikov@mm94.ru,  E.Burnaev@skoltech.ru}}
\maketitle              
\begin{abstract}
The main aim of this work is to develop and implement an automatic anomaly detection algorithm for meteorological time-series. To achieve this goal we develop an approach to constructing an ensemble of anomaly detectors in combination with adaptive threshold selection based on artificially generated anomalies. We demonstrate the efficiency of the proposed method by integrating the corresponding implementation into ``Minimax-94'' road weather information system.
   
\end{abstract}
\section{Introduction}

Effective operation of federal highways in Russia during the winter period is a very complicated and important task which reduces the number of road accidents and incidents significantly. 
Sleet, frost, low visibility are several examples of meteorological conditions on the road that increase the car crash chances. 
Preventing such conditions and reducing their consequences is a complicated problem to deal with for two main reasons. First, they require immediate measures like involving snow removal machinery, reducing the speed limits or even closing certain parts of the road. Secondly, dangerous conditions are difficult to recognize using global weather reports as they depend strongly on local road conditions like the presence of water bodies, forests, traffic intensity, etc. Very strong locality is a real challenge in this case.

An urgent need of constant monitoring road surface meteorological conditions caused development and implementation of what is called \textit{Road Weather Information Systems} (RWIS) \cite{buchanan2005road, pinet2003development, toivonen2001road}. A standard RWIS consists of three main components:  monitoring, forecasting, and decision supporting.

As in any multi-component system, there are a lot of sources of errors. In addition to the entirely autonomic and automatic mode of meteorological stations, they work in an aggressive environment with sharp temperature changes, heavy precipitation and a wide range of mechanical impacts. Breakage of sensors, external impacts, server connection errors result in providing unreliable data and creating incorrect forecasts. The consequences of errors can be different from road accidents caused by improper road service to environmental problems caused by an excessive amount of road salts poured onto the surface.

Clearly, in the case of the road information system, the mentioned problem of malfunctioning equipment and incorrect data delivery detection can be formulated as a problem of outliers/anomalies detection in the sensor data, collected by RWIS (in this work we use words ``outlier'' and ``anomaly'' interchangeably).

In this work, we propose the ensemble-based approaches for creating an outlier detector to be used for different kinds of outliers and artificial outliers generation for the optimal threshold selection. Ensemble techniques have achieved significant success in many data mining problems in recent years
\cite{da2014tweet, seni2010ensemble, salehi2016smart}. 
Although the ensemble approach for outlier detection is not as widely studied as for classification and clustering tasks, several algorithms showed considerable improvement and allowed to achieve significant advances \cite{aggarwal2017outlier, zimek2013subsampling,EnsemblesDetectors2015}. The proposed method allows coping with heterogeneous outliers caused by different factors due to its ensemble structure, as different individual algorithms in the ensemble compensate errors of each other. 

Moreover, the proposed method allows selecting a threshold for the given consensus function which is very close to the threshold, which is optimal w.r.t. $F_1$-score metric.

We conducted experiments on the data provided by ``Minimax-94'' to verify the proposed approach. We used records for five years collected with the frequency of two measurements per hour from 59 meteorological stations: we utilized data from 50 stations to train the anomaly detector and data from nine stations to test the model. To prove the general applicability of the method for multidimensional datasets without time series structure we also conducted experiments on the so-called Shuttle dataset.

The paper is organized as follows. In Section \ref{Sec2} we discuss related work. In Section \ref{proposed_method} we describe the proposed methods. We discuss our experimental setup in Section \ref{experiment_setup} and the obtained experimental results in Section \ref{experiment_results}. In Section \ref{Sec6} we draw conclusions.

\section{Related Work}
\label{Sec2}

In this section, we introduce a survey of existing outlier detection techniques, some of which are used in this paper as base learners to build an ensemble of anomaly detectors. We describe their motivations, comparative advantages, disadvantages and underlying assumptions. 

Local Outlier Factor (LOF) \cite{breunig2000lof} is a typical representative of density-based methods. It is based on computing a local density of each point and comparing it with the density of its neighborhood. Thus, the anomaly score of LOF is the ratio of the local density of this point and the local density of its nearest neighbors. The points with density lower than that of their $k$ neighbors are considered to be outliers. There are also several extensions and modifications of Local Outlier Factor algorithm. 
For example, Local Correlation Integral (LOCI) 
\cite{papadimitriou2003loci}, which uses 
Multi-Granularity  Deviation  Factor  (MDEF) as an abnormal measure. MDEF expresses how the number of data points in the neighborhood of a particular point compares with that of the points in its neighborhood.

Outlier detection methods based on statistical parametric approaches work under the assumption that data is generated from a known distribution with unknown parameters. 
Thereby, the way for detecting outliers with such models includes two steps: training and testing. Train step implies the estimation of the distribution parameters. Test step involves checking whether a new instance was generated from the probabilistic model with parameters determined on the previous step or came from another distribution. 
Usually, a class of distributions to model normal data is based on some prior assumptions and a user personal experience.
 For example, Elliptic Envelope uses multivariate Gaussian distribution assumption and robust covariance matrix estimation by MinCovarianceDet estimator \cite{rousseeuw1999fast}.

Outlier detection methods based on regression analysis have been extensively used for time series data. The procedure of regression-based outlier detection is generally the following: a regression model is fitted to the data during the training phase, during the test phase the fitted model is applied to the new data instances and the abnormality of each instance is evaluated according to the obtained residuals. If the instance is located far from the regression line, we mark it as an outlier. We can use regression conformal confidence measures to estimate an abnormality threshold \cite{VovkConformal2014, ConformalKRR2016, ConformalDR}. Note that we can efficiently utilize the same methodology for conformal measures construction to define non-parametric anomaly detectors in time-series data \cite{ConformalMartingales2017, ConformalAD2015, kNN2017}.
    
One-class Support Vector Machine (OCSVM) \cite{scholkopf2000support} is an outlier detection modification of Support Vector Machine (SVM). While in supervised tasks SVM aims to find a maximal margin hyperplane separating two classes, OCSVM aims to separate training data instances from the origin. In \cite{ModelSelection2015} they consider approaches to OCSVM model selection. Generalization of OCSVM taking into account privileged information is provided in \cite{OCSVM2016, OCSVM2018}.

Isolation Forest (IForest) \cite{liu2008isolation} is a successful modification of the standard decision tree-based algorithm for outlier detection. 
It randomly selects a feature and then randomly selects a split value from the interval between the minimum and the maximum values of the chosen parameter. 
The process continues until each leaf of the tree is assigned to only one observation from the dataset. 
The primary assumption of the algorithm is that anomalous observations can be separated during the first steps of this process because the algorithm requires fewer conditions to isolate the anomalies from normal instances. 
Normal observations, on the contrary, require more conditions to be separated from each other. The resulting anomalous score equals the length of the path from the root to the leaf.

If at least partial labeling of anomalies is known, we can use approaches to imbalanced classification for the anomaly detector construction, see \cite{Imbalanced2019, Imbalanced2015}.

\section{Proposed Method}
\label{proposed_method}

The meteorological time series dataset explored in this work is notable for the diverse nature of occurring outliers. Anomalies in the given dataset are caused by many different factors like sensor malfunction, some external events like a bird sitting on the station, server connection errors, etc. An individual algorithm can hardly cope with the whole range of outliers, due to strong underlying assumptions. Hence, exploiting ensemble analysis seems to be an appropriate choice.
In this section, we propose an ensemble outlier detection method with artificial anomalies generation for selecting an optimal threshold. We affirm that the proposed approach has several advantages over the existing methods, especially in the case of outlier detection in meteorological time series.  

Most algorithms require either defining an exact threshold for outlier scores or percentage of outliers in the given data to convert the scores into binary labels.  The proposed method is aimed at solving the mentioned problem of selecting an optimal threshold. The first step of the algorithm includes dividing the data into two samples: for training ensemble base models and for selecting the threshold. The next and essential step is generating some amount of artificial anomalies and adding them to the second part of the data. The pseudo code for the proposed method is given in Algorithm \ref{alg:proposed_method}.

\begin{algorithm}[t!] 
\caption{Threshold Selection Using Artificial Anomalies} 
\label{alg:proposed_method}
\begin{flushleft}
        \textbf{Input:} \\
        $D$ -- given dataset; 
        $E$ -- set of base ensemble models; \\
        $G$ -- distribution to generate anomalies from;
        $N$ -- number of outliers;\\ 
        
        \textbf{Output:} \\
        $\Tilde{E}$ -- trained ensemble base methods; 
        $T$ -- threshold
        \end{flushleft}
\begin{algorithmic}[1]
    \STATE Divide $D$ into $D_{train}$ for training and $D_{thresh}$ for selecting threshold 
    
    \FOR{$i$ in range $(1, N)$}
        \STATE Generate an outlier $o \sim G$
        \STATE $D_{thresh} = D_{thresh} \bigcup \lbrace o \rbrace$
    \ENDFOR
    
    \STATE $S_{thresh} = \emptyset$ -- scores obtained on $D_{thresh}$
    
    \FOR{algorithm $e$ in $E$}
        \STATE Train $e$ on $D_{train}$
        \STATE Obtain $s_e$ -- output score of $e$ on $D_{thresh}$
        \STATE $S_{thresh} = S_{thresh} \bigcup \lbrace s_e \rbrace$
    \ENDFOR
    \STATE Combine scores $S_{thresh}$ and get $s_{final}$
    \STATE Select threshold $T$ for $s_{final}$, which optimizes the quality metric
\end{algorithmic}
\end{algorithm}

\section{Experiment Setup}
\label{experiment_setup}

\subsection{Dataset Description}

Company ``Minimax-94'' provided the primary dataset used in this work. ``Mini\-max-94'' is the research and production company working in the road industry and specializing in the creation of intelligent transport meteorological control systems. Archived data received from RWIS and provided for the research contains records for the period from 2012 to  2017. The stations aggregate data and send it to the data storage with a frequency of approximately two measurements per hour. For the train part, we selected 50 stations covering all main Russian climate zones. Then we divided them into two subsets.

The first subset of stations consists of 35 stations ($\sim 2 \cdot 10^6$ time ticks). There are two different versions of data from these stations: initial with noise, and clean data. The cleaning process was applied only to road surface temperature; other components remain in their original form. We removed each anomaly point with a rather big neighborhood of length equal to several days.
        
The second part of the training sample consists of only clean records from the rest 15 meteorological stations ($\sim 1 \cdot 10^6$ time ticks). It is used for creating artificial outliers and the threshold selection.

The test contains records from nine stations with labeled data; we marked each anomaly point with its 2-3 hour local neighborhood, i.e., we labeled all data points from anomaly's vicinity as outliers. In total test data contains about 
    $8 \cdot 10^5$ records, and $\sim1\%$ of which we marked as outliers.  

\begin{figure}
    \centering
    \includegraphics[width=0.7\textwidth]{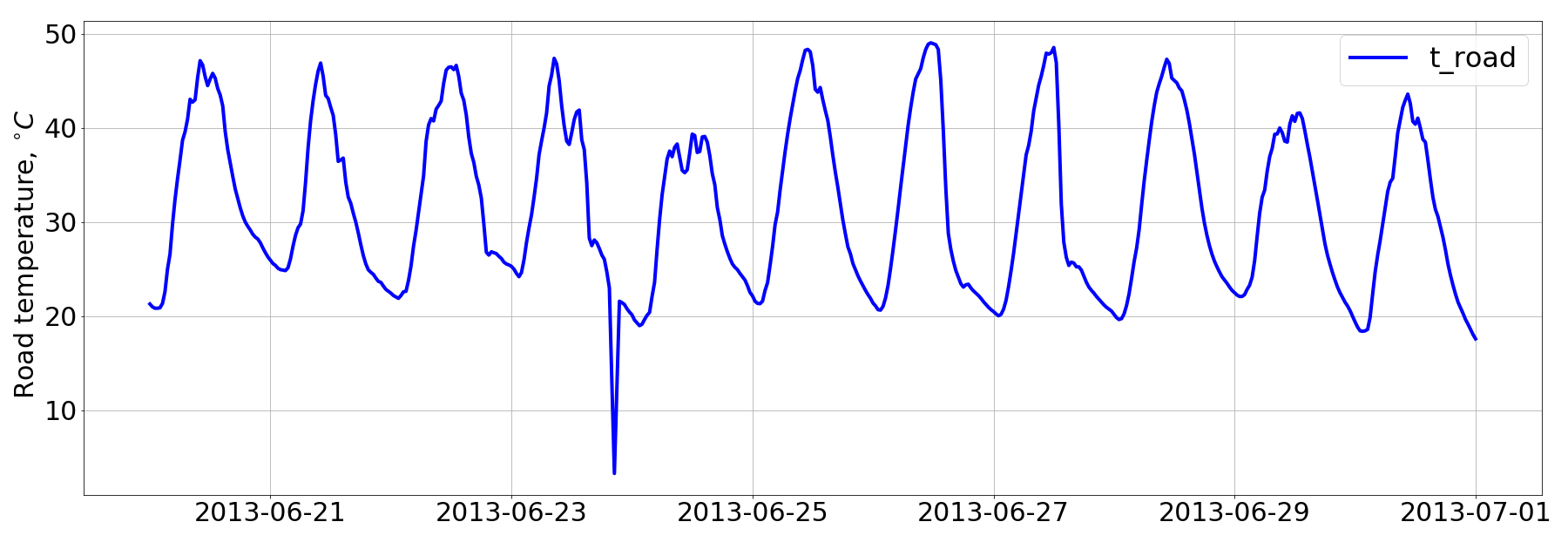}
    \caption{Example of a single outlier}
    \label{fig:short_outlier}
    \end{figure}

\begin{figure}
    \centering
    \includegraphics[width=0.7\textwidth]{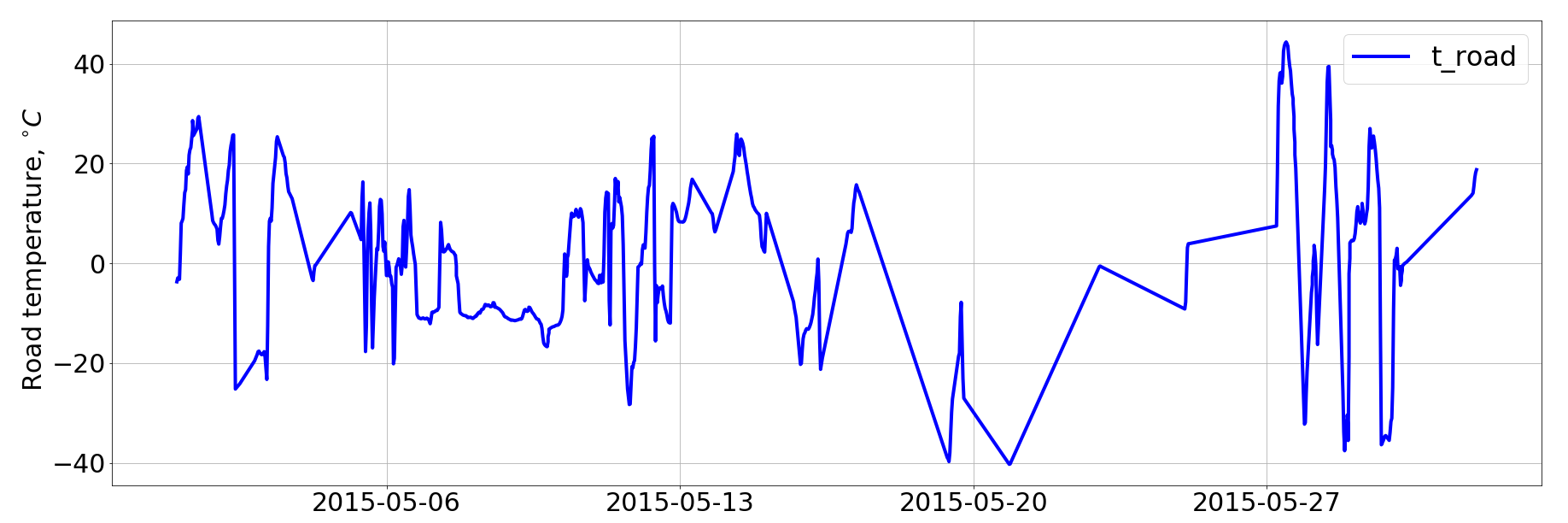}
    \caption{Example of long-term malfunction}
    \label{fig:long_outlier}
    \end{figure}

We can roughly categorize typical outliers behavior into two categories: short-term anomalies usually caused by some external impact, like exhaust pipe directed at the sensor (Figure \ref{fig:short_outlier}) and long-term anomalies caused by serious station malfunction (Figure \ref{fig:long_outlier}).

To prepare data we conducted the following steps. If the time gap between two nearest records from RWIS is longer than two hours, then the time series is divided into two parts. We remove patterns with duration less than 12 hours from the sample. Finally, we linearly interpolate the data.


Also, we consider another dataset to test the proposed approach --- the Shuttle dataset. It has been generated to classify which type of control to use during the landing of the spacecraft depending on the external conditions. The dataset contains 58000 instances, described by nine integer attributes and divided into seven classes: 1: Rad Flow, 2: Fpv Close, 3: Fpv Open, 4: High, 5: Bypass, 6: Bpv Close, 7: Bpv Open. We attribute classes 1 and 4 to normal instances while assigning the rest five categories to outliers. Two normal classes constitute about 94\% of the sample, so the number of outliers is 6\%.

For the test part, we selected 25\% of the dataset. Remaining 75\% were split into two parts: 2/3 is used to train and validate outlier detection algorithms and consists of both normal and abnormal classes, the other 1/3 is used to select an optimal threshold for an ensemble of algorithms using artificially generated outliers and includes only instances of the normal classes.

\subsection{Anomaly Generation}

\begin{figure}[t]
     \centering
    \includegraphics[width=0.7\textwidth]{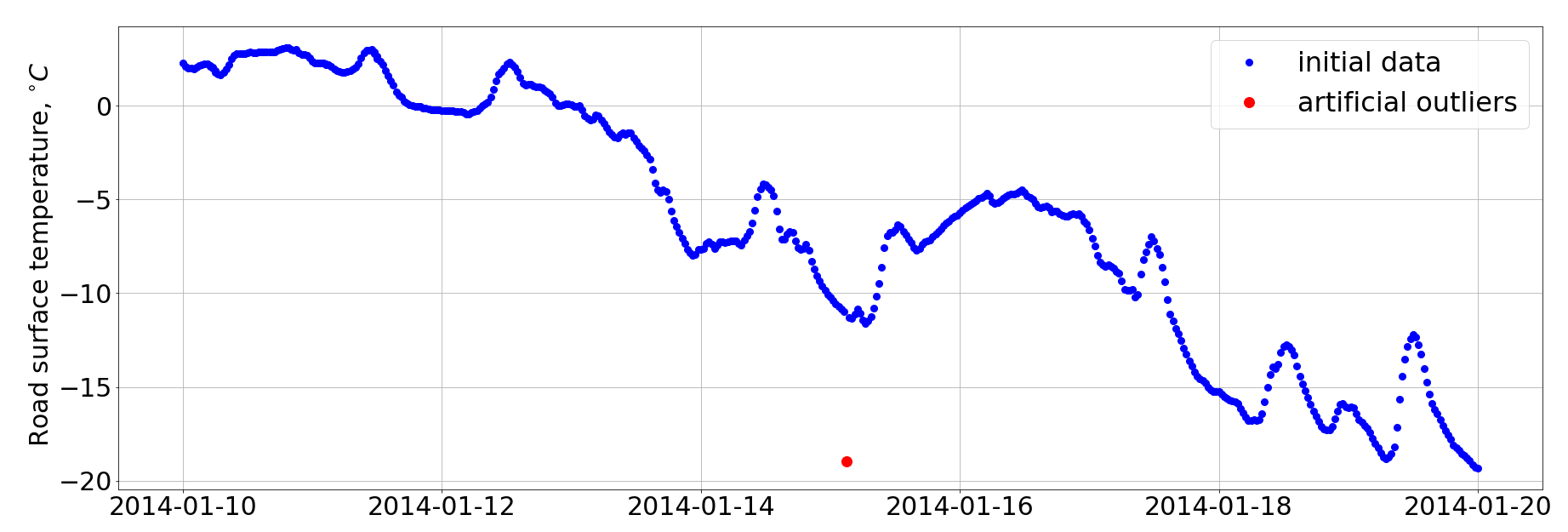}
    \caption{Example of artificially generated single outlier}
    \label{fig:single_outlier_generated}
    \end{figure}

\begin{figure}[t]
    \centering
    \includegraphics[width=0.7\textwidth]{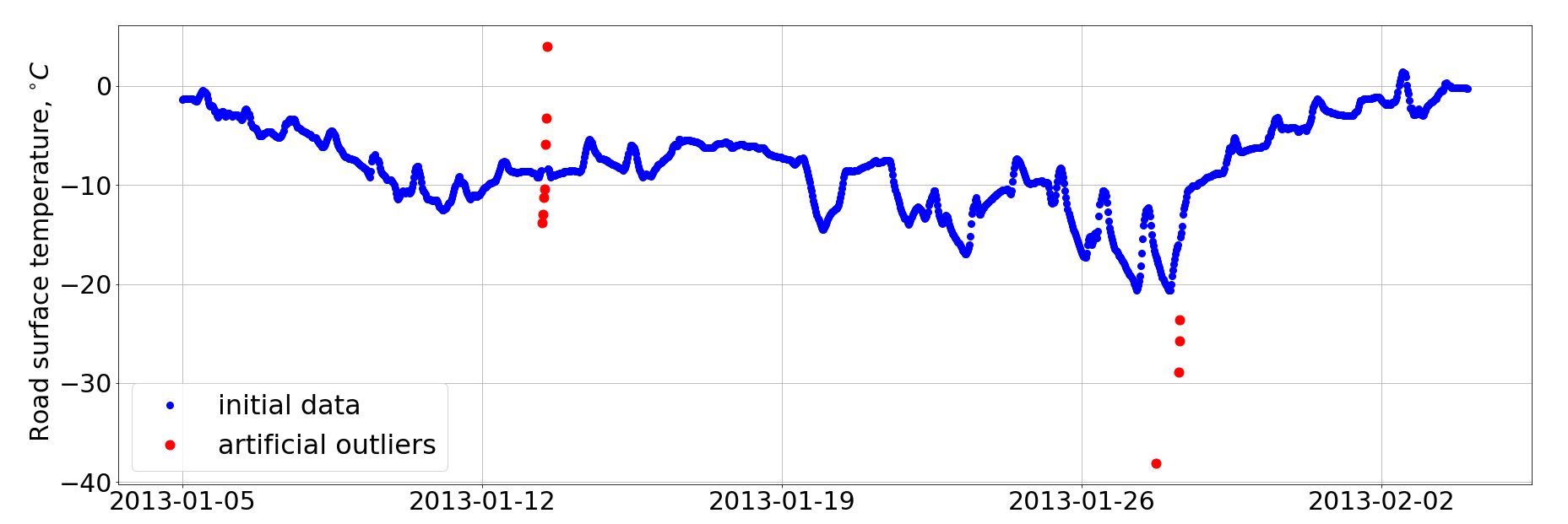}
    \caption{Example of artificially generated short-term anomalous series}
    \label{fig:short_outlier_generated}
    \end{figure}

\begin{figure}[t]
    \centering
    \includegraphics[width=0.7\textwidth]{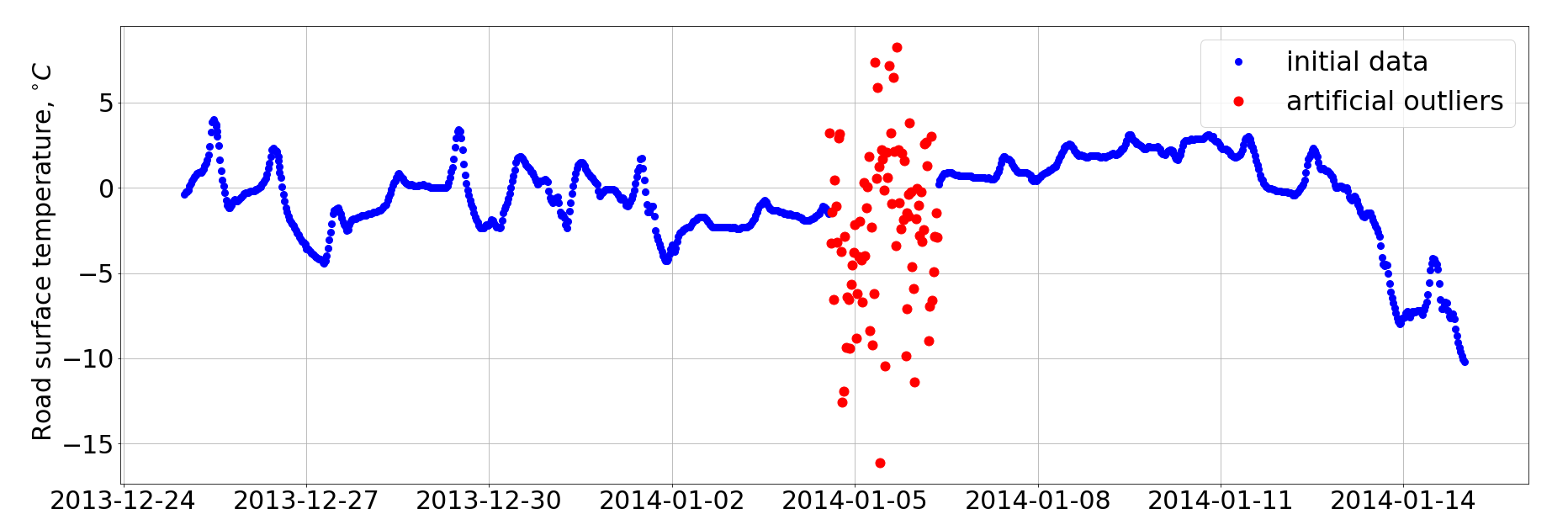}
    \caption{Example of artificially generated long-term sensor malfunction}
    \label{fig:long_outlier_generated}
    \end{figure}

Since real outliers probably come from different distributions and have different nature, we model the most frequent examples of outliers occurring in the given time series. They are single outliers (see Algorithm~\ref{alg:single_outlier_generation} and Figure~\ref{fig:single_outlier_generated}), short-term outliers (see Algorithm~\ref{alg:short_outlier_generation} and Figure~\ref{fig:short_outlier_generated}) and long-term sensor malfunctions (see Algorithm \ref{alg:long_outlier_generation} and Figure \ref{fig:long_outlier_generated}). For each station we generate  30 single outliers, 20 short-term and 3 long-term anomalies. Artificial labels are generated based on real-life data which was labeled by experts. During the labeling, the experts eliminated the sudden weather changes and so these events were not marked as sensors malfunctions.

    \begin{algorithm}[t] 
    \caption{Single Outliers Generation} 
    \label{alg:single_outlier_generation}
    \begin{flushleft}
            \textbf{Input:} \\
            $S$ -- list of meteorological stations; \\
            $t_{road}$ -- road surface temperature time series; \\
            $N = 30$ -- number of outliers per each station in $S$; \\ $a_{low} = 2,~~a_{up} = 5$ -- lower and upper boundaries of the uniform distribution \\
            \textbf{Output:} $\Tilde{t}_{road}$ -- road surface temperature time series along with artificially generated anomalies
    \end{flushleft}
    \begin{algorithmic}[1] 
        \STATE $\Tilde{t}_{road} = t_{road}$
        \FOR {each station $s$ in $S$}
            \STATE Randomly select $N$ time stamps from
            $\Tilde{t}_{road}[s, :]$
            \FOR {each selected time stamp $t$}
                \STATE Generate perturbation $p \sim U(a_{low}, a_{up})$
                \STATE Randomly select $sign$ from $\lbrace -1, 1 \rbrace$ 
                \STATE $\Tilde{t}_{road}[s, t] = t_{road}[s, t] + sign * p$ 
            \ENDFOR
        \ENDFOR
    \end{algorithmic}
    \end{algorithm}

    \begin{algorithm}[t]
    \caption{Short-Term Anomaly Generation}
    \label{alg:short_outlier_generation}
    \begin{flushleft}
            \textbf{Input:} \\
            $t_{road}$ -- road surface temperature time series;
            $S$ -- list of meteorological stations; \\
            $N = 20$ -- number of anomalies per each station in $S$; \\
            $\lambda = 2$ -- rate parameter of exponential distribution; \\
            $d_{low} = 3,~~d_{up} = 12$ -- lower and upper boundaries of anomalous duration \\
            \textbf{Output:} $\Tilde{t}_{road}$ -- time series along with artificially generated anomalies
    \end{flushleft}
    \begin{algorithmic}[1] 
        \STATE $\Tilde{t}_{road} = t_{road}$
        \FOR {each station $s$ in $S$}
            \STATE Randomly select $N$ time stamps from 
            $\Tilde{t}_{road}[s, :]$
            \FOR {each selected time stamp $t$}
                \STATE $d \sim RandInteger(d_{low}, d_{up})$
                \STATE Randomly select $sign$ from $\lbrace -1, 1 \rbrace$ 
                \STATE Perturbation array $p = zeros(d)$
                \FOR {i in range(2, d)}
                    \STATE $p[i] \sim Exp(\lambda)$ \\
                    \STATE $p[i] = p[i] + p[i-1]$
                \ENDFOR
                \STATE $\Tilde{t}_{road}[s, t : (t + d)] = t_{road}[s, t: (t + d)] + sign * p$ 
            \ENDFOR
        \ENDFOR
    \end{algorithmic}
    \end{algorithm}

    \begin{algorithm}[t]
    \caption{Long-Term Anomaly Generation}
    \label{alg:long_outlier_generation}
    \begin{flushleft}
            \textbf{Input:} \\
            $t_{road}$ -- road surface temperature time series; \\
            $S$ -- list of meteorological stations; \\
            $N = 3$ -- number of anomalies per each station in $S$; \\
            $a_{low} = 30,~~a_{up} = 200$ -- lower and upper boundaries of multiplier; \\
            $d_{low} = 30,~~d_{up} = 200$ -- lower and upper boundaries of anomalous duration; \\
            $\mu = 0,~~\sigma = 5$ -- mean and standard deviation of perturbation \\
            \textbf{Output:} $\Tilde{t}_{road}$ -- time series along with artificially generated anomalies
    \end{flushleft}
    \begin{algorithmic}[1] 
        \STATE $\Tilde{t}_{road} = t_{road}$
        \FOR {each station $s$ in $S$}
            \STATE Randomly select $N$ time stamps from 
            $\Tilde{t}_{road}[s, :]$
            \FOR {each selected time stamp $t$}
                \STATE $d \sim RandInteger(d_{low}, d_{up})$
                \STATE Randomly select multiplier $mult$ from $Uniform(m_l, m_u)$ 
                \STATE Perturbation array $p = zeros(d)$
                \FOR {i in range(1, d)}
                    \STATE $p[i] \sim N(\mu, \sigma)$ \\
                \ENDFOR
                \STATE $\Tilde{t}_{road}[s, t : t + d] = mult * t_{road}[s, t: (t + d)] + p$ 
            \ENDFOR
        \ENDFOR
    \end{algorithmic}
    \end{algorithm}

The algorithm for generating artificial outliers for the Shuttle dataset consists of three steps.  Transform the train data using Principal Component Analysis (PCA) decomposition, consider only 2 components corresponding to the largest singular values. Randomly generate 450 instances with the first component from $Uniform(-10000, 10000)$ and the second component equal to zero. Randomly generate 450 instances with the first component equal to zero and the second component from $Uniform(-5000, 5000)$. Conduct inverse PCA transformation.

\subsection{Evaluation}
The meteorological test sample is labeled as follows: for each anomaly point we mark its 2-3 hour neighborhood as anomalous.

The developed quality metric is an extension of simple $F_1$-score to the considered case. The standard $F_1$-the score is defined as

\begin{equation}
F_1 = 2 \cdot \frac{recall \times precision}{recall + precision},
\end{equation}
where recall is the fraction of true positive instances over the number of all real positive instances,
while precision is the fraction of true positive points among all the instances labeled as positive. The modified version of the quality metric is the following. Recall is the fraction of instances labeled as outliers (by the algorithm) in the vicinity of which there is at least one instance marked as an outlier (real label) to the total amount of positive instances (real labels).
Precision is the fraction of instances labeled as outliers (real label) in the vicinity of which there is at least one instance marked as an outlier (by the algorithm) to the total amount of positive instances (by the algorithm).

\subsection{Base Learners for Anomaly Detection}

We consider several outlier detection approaches to test the effectiveness of the proposed method for threshold selection: single regression models, model averaging, feature bagging ensemble. All of them except for Multi Layer Perceptron (MLP) use 35 clean stations as a training set. Forecasting based anomaly detection approach predicts target value 30 minutes ahead.

As features, we used air, road surface and subsurface temperatures, pressure, and humidity for several previous hours; besides that we used differences between the first six lags, solar azimuth, and altitude angles, road id, longitude and latitude of the station, sine and cosine of the hour, day of the year and month.

We tested several algorithms: XGBoost Regression  \cite{Chen:2016:XST:2939672.2939785}, MLP (2 hidden layers with 64 and 16 neurons on the first and the second layers relatively, rectified linear unit (ReLU) activation function), XGBoost Air, One-class SVM (OCSVM), Elliptic Envelope (Ell. Env.), Local Outlier Factor (LOF), Isolation Forest (IForest), 

In case of the Shuttle dataset model averaging includes the following algorithms: Local Outlier Factor, Isolation Forest, Elliptic Envelope, One-class SVM, Ridge (we use the first feature as a target variable, and others as independent variables).

We created two feature bagging based ensembles on the base of Ridge Regression, Elliptic Envelope for the meteorological dataset. We implemented feature bagging in combination with Ridge Regression, Elliptic Envelope, Local Outlier Factor and  One-Class SVM for the Shuttle dataset.

Ensembles built for the meteorological dataset include some minor modifications concerning the number of selected features, i.e., the amount of features is generated from an interval that is different from the initial $ [\floor*{d/2}, d-1] $ interval, where $d$ is the complete number of features: 1) Ridge Regression --- 20 models in the ensemble, number of features is a random integer from $\floor*{d/6}$ to $\floor*{d/2}$;
2) Elliptic Envelope --- 10 models in the ensemble, number of features for each model is a random integer from $2$ to $\floor*{(2 \cdot d)/3}$.
    
Ensembles built for the Shuttle dataset completely repeat the described method. Shuttle dataset ensembles consist of 20 models each, and the number of features is a random integer 
from $\floor*{d/2}$ to $ d-1$ interval.

\subsection{Combination of Scores}
\label{bunch_of_models}

To build a combination of predictions, we normalize predictions to [0, 1] and try three approaches: simple averaging, averaging with weights equal to Pearson correlation between scores on the artificial data sample and the vector with labels, and finally training logistic regression with model scores as an input feature vector and artificial data as labels.

\section{Experimental Results}
\label{experiment_results}
\begin{figure}[t]
    \centering
    \includegraphics[width=0.7\textwidth]{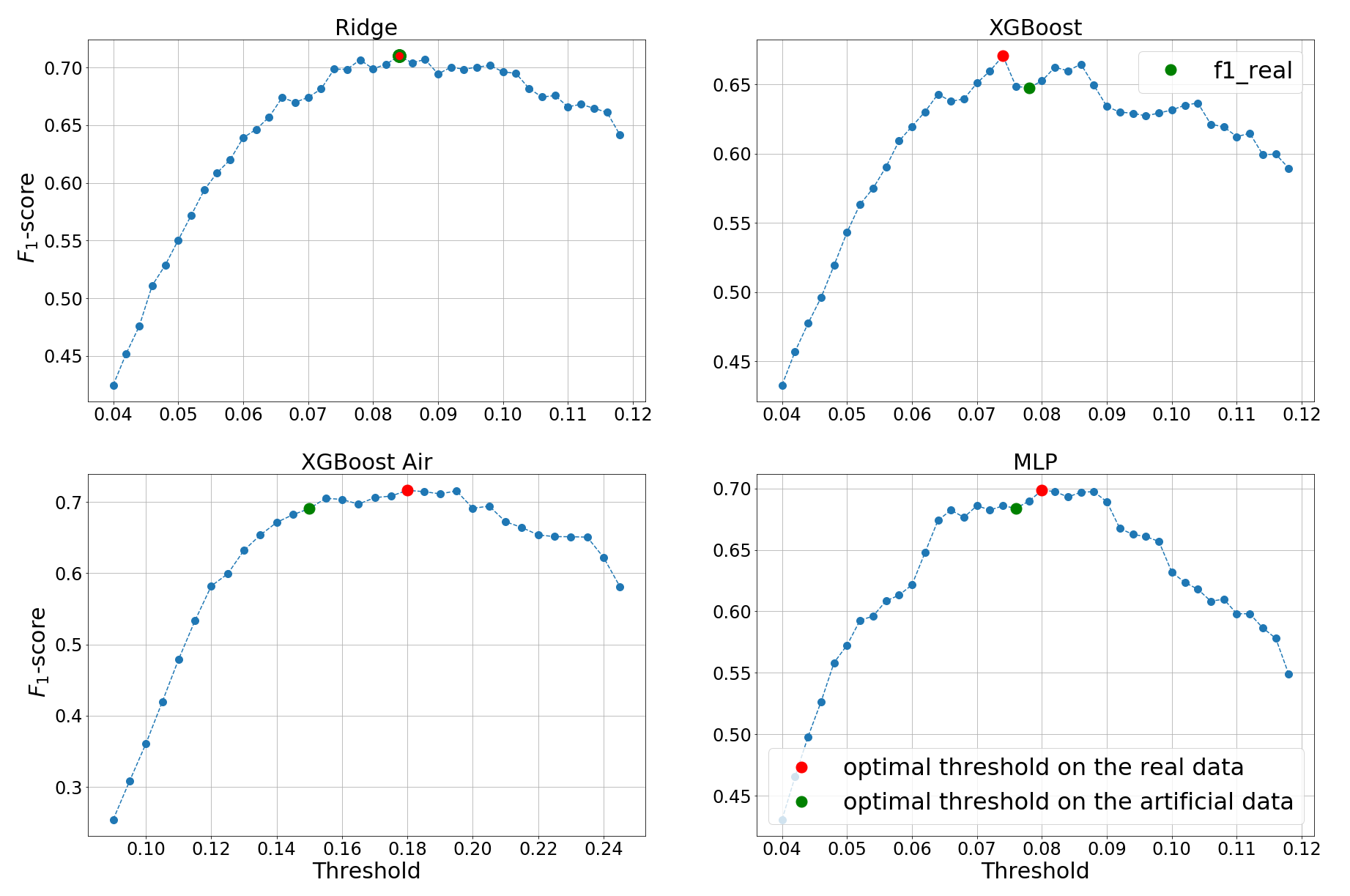}
    \caption{Threshold selection using artificial anomalies for the meteorological dataset}
    \label{fig:individual_results}
\end{figure}

The proposed threshold selection method shows excellent performance in case of applying individual regression algorithms to the meteorological dataset. 
The main advantage of the method is a rather precise estimation of the $F_1$-optimal threshold obtained on the artificial data. 
As can be seen in Figure \ref{fig:individual_results} the threshold selected using the generated data is very close to the threshold optimizing real $F_1$-score on the actual data.
    
As described in Section \ref{experiment_setup}, we examined three different ensembles for the meteorological dataset: model-centered model averaging and two data-centered ensembles based on feature bagging with Ridge and Elliptic Envelope base algorithms (denoted as ``Ridge FB'', ``Ell.Env. FB''). For each algorithm different combination functions have been used: an average of the scores linearly transformed into [0, 1] interval (``LT''), a weighted average of the linearly transformed scores (``WLT'') and a Logistic Regression (``LogReg''). 

We compared the ensemble approach with individual outlier detectors on the meteorological dataset. In this experiment, all individual algorithm results serve as a baseline for the ensemble approach. We provided the results of the comparison in Table \ref{tab:results_meteo}. As can be observed from the table, almost all ensemble techniques outperformed individual algorithms and, what is more critical, outperformed all included base ensemble models. In case of the meteorological data Logistic Regression consensus function turned out to show the best performance, although it is a kind of ``risky'' choice as it tends to overfit and as a result is somewhat unstable, which we can observe when using Elliptic Envelope. The similar situation happens with the Shuttle dataset.

\begin{table}[t]
    \centering
    \caption{Results: Meteorological Dataset}
\label{tab:results_meteo}
    \begin{tabular}{|c|c|c|c|c|c|c|c|c|}
    \hline
         Algorithm & Ridge & XGBoost & MLP & XGBoost Air & LOF & Ell.Env & OCSVM & IForest  \\
        \hline
         Recall  & 0.668 & 0.640 & 0.637 & 0.662 & 0.513 & 0.719 & 0.904 & 0.395\\
        \hline
        Precision & 0.758 & 0.655 & 0.738 & 0.722 & 0.502 & 0.482 & 0.130 & 0.252\\
        \hline
        $F_1$-score & 0.710 & 0.647 & 0.684 & 0.691 & 0.507 & 0.577 & 0.227 & 0.308\\
        \hline
    \end{tabular}
        
    \begin{tabular}{|c|c|c|c|c|c|c|c|c|c|}
    \hline
        Ensemble & \multicolumn{3}{c|}{Model Averaging}  & \multicolumn{3}{c|}{Ridge FB} & \multicolumn{3}{c|}{Ell.Env FB}\\
        \hline
        Combination & LT & WLT & LogReg & LT & WLT & LogReg & LT & WLT & LogReg \\
        \hline
        Recall & 0.701 & 0.668 & 0.753 & 0.671 & 0.685 & 0.690 & 0.691 & 0.696 & 0.76\\ 
        \hline
        Precision & 0.750 & 0.802 & 0.797 & 0.789 & 0.731 & 0.829 & 0.739 & 0.708 & 0.630\\
        \hline
        $F_1$-score & 0.725 & 0.729 & 0.774 & 0.725 & 0.707 & 0.753 & 0.714 & 0.701 & 0.670\\
        \hline

    \end{tabular}
\end{table}

Table \ref{tab:shuttle_dataset} shows that ensembling techniques improve the results of the individual algorithms significantly almost in all cases for the Shuttle Dataset. The most considerable improvement of $F_1$-score has been achieved by One-class SVM, which increased it from 0.272 to 0.999. As a result, feature bagged OCSVM achieved the best result for the Shuttle dataset.

\begin{table}[t]
    \centering
    \caption{Results: Shuttle Dataset}
\label{tab:shuttle_dataset}
    \begin{tabular}{|c|c|c|c|c|c|}
    \hline
        Algorithm & Ridge & LOF & Ell.Env & OCSVM & IForest \\
        \hline
        Recall & 0.955 & 1.00 & 0.704 & 0.461 & 0.799\\
        \hline
        Precision & 0.767 & 0.512 & 0.680 & 0.193 & 0.791 \\
        \hline
        $F_1$-score & 0.851 & 0.677 & 0.692 & 0.272 & 0.795\\
        \hline
    \end{tabular}
    \begin{tabular}{|c|c|c|c|c|c|c|c|c|c|}
    \hline
        Ensemble & \multicolumn{3}{c|}{Model Averaging}  & \multicolumn{3}{c|}{Ridge FB} & \multicolumn{3}{c|}{Ell.Env FB}\\
        \hline
        Combination & LT & WLT & LogReg & LT & WLT & LogReg & LT & WLT & LogReg\\
        Recall & 0.957 & 0.924 & 0.756 & 0.956 & 0.953 & 0.681 & 0.625 & 0.567 & 0.476\\
        \hline
        Precision & 0.979 & 0.986 & 0.815 & 0.871 & 0.938 & 0.801 & 0.937 & 0.930 & 0.915\\
        \hline
        $F_1$-score & 0.968 & 0.953 & 0.784 & 0.911 & 0.946 & 0.736 & 0.750 & 0.704 & 0.626\\
        \hline
    \end{tabular}
    \\
    
    \begin{tabular}{|c|c|c|c|c|c|c|}
    \hline
        Ensemble &  \multicolumn{3}{c|}{LOF FB} & \multicolumn{3}{c|}{OCSVM FB}\\
        \hline
        Combination & LT & WLT & LogReg & LT & WLT & LogReg\\
        \hline
        Recall & 0.587 & 0.577 & 0.311 & 0.981 & 0.999 & 0.999\\
        \hline
        Precision & 0.891 & 0.850 & 0.882 & 0.992 & 0.998 & 0.998\\
        \hline
        $F_1$-score  & 0.704 & 0.688 & 0.460 & 987 & 0.999 & 0.999\\
        \hline
    \end{tabular}
    
\end{table}


\section{Conclusions}
\label{Sec6}

In this paper, we proposed a new method for outlier detection in the RWIS data. The method allows to detect anomalies of heterogeneous nature efficiently, makes the selection of the decision rule less subjective and also turns out to be applicable not only to the RWIS data but also to more general cases of multidimensional datasets. 

The results of the experiments show us that ensembles tend to outperform individual algorithms for both datasets. Ensembles tested on the meteorological data show better performance than all compared individual methods, which proves the assumption about benefits of ensemble-based methods for detecting anomalies in the RWIS data. There is no such strong tendency for the Shuttle data, for example, single Ridge Regression or Isolation Forest obtain higher $F_1$-scores than LOF feature bagging ensemble. However, even for the Shuttle dataset, the combination of different ensembles received higher $F_1$-score than any individual algorithm included in the combination.

The comparison of different combination functions showed that although linear transformation seems to perform better than weighted linear transformation, there is no significant difference between these two methods and choosing the best one largely depends on the specific setting of the problem, ensemble technique and included individual algorithms. Ensembles achieve the best results on both datasets with Logistic Regression. However, this combination method can easily overfit especially in case of the Shuttle dataset and is considered to be the most unstable, as providing the best and the worst results depending on the ensemble technique.

The proposed threshold selection technique showed significant performance for both ensemble approaches and individual algorithms. We can observe that the thresholds optimizing $F_1$-score for the artificial data are very close to the ones for the real data. 

An ensemble with the best results, i.e., a model averaging (Section \ref{bunch_of_models}) with Logistic Regression consensus function, has been chosen as the principal method for outlier detection in ``Minimax-94'' road information system. It was implemented in Python as a separate module and integrated into the company system in the test mode. According to the experiments conducted on the archive records the selected method allows to detect 75\% of occurring anomalies with 20\% false alarm rate, which is sufficient for ``Minimax-94''. The future work assumes aggregation of statistics obtained in online mode during the test period and incorporation of the module into the working cycle of the ``Minimax-94'' system. To perform online aggregation, we are going to use long-term aggregation strategies \cite{AggregationLongTerm} along with approaches to model quasi-periodic data \cite{QuasiPeriodic} and extraction of trends in the presence of non-stationary noise with long tails \cite{Degradation2016, FBM2016}. Approaches to multidimensional time-series prediction \cite{MF2018} and multichannel anomaly detection \cite{Multichannel2017} will allow detecting complex anomalies related to change of dependencies between time-series components.

We also would like to note that the resulting approach is highly dependent on the algorithm of synthetic data generation. We selected this technique based on certain domain knowledge. In order to adapt the technique for any new problem we have to find out what are the most common patterns of anomalous data appearing due to malfunctioning sensors. If due to some reason behavior of the sensors changes and malfunctioning sensors start to produce other patterns of anomalous data or some drastic weather change happens, then there is a chance that our algorithm fails to detect anomalies. However, this is a common problem for all machine learning algorithms.

%
%
%
\bibliographystyle{splncs04}
\bibliography{references.bib}

\begin{thebibliography}{10}
\providecommand{\url}[1]{\texttt{#1}}
\providecommand{\urlprefix}{URL }
\providecommand{\doi}[1]{https://doi.org/#1}

\bibitem{aggarwal2017outlier}
Aggarwal, C.C., Sathe, S.: Outlier ensembles: an introduction. Springer (2017)

\bibitem{EnsemblesDetectors2015}
Artemov, A., Burnaev, E.: Ensembles of detectors for online detection of
  transient changes. In: Eighth International Conference on Machine Vision
  (ICMV 2015), 98751Z (8 December 2015). Proc. SPIE, vol.~9875 (2015)

\bibitem{Degradation2016}
Artemov, A., Burnaev, E.: Detecting performance degradation of
  software-intensive systems in the presence of trends and long-range
  dependence. In: 2016 IEEE 16th Int. Conf. on Data Mining Workshops (ICDMW).
  pp. 29--36 (2016). \doi{10.1109/ICDMW.2016.0013}

\bibitem{FBM2016}
Artemov, A., Burnaev, E.: Optimal estimation of a signal perturbed by a
  fractional brownian noise. Theory of Probability \& Its Applications
  \textbf{60}(1),  126--134 (2016)

\bibitem{QuasiPeriodic}
Artemov, A., Burnaev, E., Lokot, A.: Nonparametric decomposition of
  quasi-periodic time series for change-point detection. In: Eighth
  International Conference on Machine Vision (ICMV 2015), 987520 (8 December
  2015). Proc. SPIE, vol.~9875 (2015)

\bibitem{breunig2000lof}
Breunig, M.M., Kriegel, H.P., Ng, R.T., Sander, J.: Lof: Identifying
  density-based local outliers. SIGMOD Rec.  \textbf{29}(2),  93--104 (2000).
  \doi{10.1145/335191.335388}, \url{http://doi.acm.org/10.1145/335191.335388}

\bibitem{buchanan2005road}
Buchanan, F., Gwartz, S.: Road weather information systems at the ministry of
  transportation, ontario. In: 2005 Annual Conference of the Transportation
  Association of Canada (2005)

\bibitem{Imbalanced2015}
Burnaev, E., Erofeev, P., Papanov, A.: Influence of resampling on accuracy of
  imbalanced classification. In: Eighth International Conference on Machine
  Vision (ICMV 2015), 987521 (8 December 2015). Proc. SPIE, vol.~9875 (2015)

\bibitem{ModelSelection2015}
Burnaev, E., Erofeev, P., Smolyakov, D.: Model selection for anomaly detection.
  In: Proc. SPIE. vol.~9875, pp. 9875 -- 9875 -- 6 (2015).
  \doi{10.1117/12.2228794}, \url{https://doi.org/10.1117/12.2228794}

\bibitem{ConformalKRR2016}
Burnaev, E., Nazarov, I.: Conformalized kernel ridge regression. In: 2016 15th
  IEEE International Conference on Machine Learning and Applications (ICMLA).
  pp. 45--52 (Dec 2016). \doi{10.1109/ICMLA.2016.0017}

\bibitem{OCSVM2016}
Burnaev, E., Smolyakov, D.: One-class svm with privileged information and its
  application to malware detection. In: 2016 IEEE 16th International Conference
  on Data Mining Workshops (ICDMW). pp. 273--280 (Dec 2016).
  \doi{10.1109/ICDMW.2016.0046}

\bibitem{VovkConformal2014}
Burnaev, E., Vovk, V.: Efficiency of conformalized ridge regression. In:
  Balcan, M.F., Feldman, V., Szepesvári, C. (eds.) Proceedings of The 27th
  Conference on Learning Theory. Proceedings of Machine Learning Research,
  vol.~35, pp. 605--622. PMLR, Barcelona, Spain (13--15 Jun 2014),
  \url{http://proceedings.mlr.press/v35/burnaev14.html}

\bibitem{Multichannel2017}
Burnaev, E.V., Golubev, G.K.: On one problem in multichannel signal detection.
  Problems of Information Transmission  \textbf{53}(4),  368--380 (Oct 2017).
  \doi{10.1134/S0032946017040056},
  \url{https://doi.org/10.1134/S0032946017040056}

\bibitem{Chen:2016:XST:2939672.2939785}
Chen, T., Guestrin, C.: {XGBoost}: A scalable tree boosting system. In:
  Proceedings of the 22nd ACM SIGKDD International Conference on Knowledge
  Discovery and Data Mining. pp. 785--794. KDD '16, ACM, New York, NY, USA
  (2016). \doi{10.1145/2939672.2939785},
  \url{http://doi.acm.org/10.1145/2939672.2939785}

\bibitem{da2014tweet}
Da~Silva, N.F., Hruschka, E.R., Hruschka~Jr, E.R.: Tweet sentiment analysis
  with classifier ensembles. Decision Support Systems  \textbf{66},  170--179
  (2014)

\bibitem{kNN2017}
Ishimtsev, V., Bernstein, A., Burnaev, E., Nazarov, I.: Conformal k-nn anomaly
  detector for univariate data streams. In: Gammerman, A., Vovk, V., Luo, Z.,
  Papadopoulos, H. (eds.) Proceedings of the Sixth Workshop on Conformal and
  Probabilistic Prediction and Applications. Proceedings of Machine Learning
  Research, vol.~60, pp. 213--227. PMLR, Stockholm, Sweden (13--16 Jun 2017),
  \url{http://proceedings.mlr.press/v60/ishimtsev17a.html}

\bibitem{AggregationLongTerm}
Korotin, A., V'yugin, V., Burnaev, E.: Aggregating strategies for long-term
  forecasting. In: Gammerman, A., Vovk, V., Luo, Z., Smirnov, E., Peeters, R.
  (eds.) Proceedings of the Seventh Workshop on Conformal and Probabilistic
  Prediction and Applications. Proceedings of Machine Learning Research,
  vol.~91, pp. 63--82. PMLR (11--13 Jun 2018),
  \url{http://proceedings.mlr.press/v91/korotin18a.html}

\bibitem{ConformalDR}
Kuleshov, A., Bernstein, A., Burnaev, E.: Conformal prediction in manifold
  learning. In: Gammerman, A., Vovk, V., Luo, Z., Smirnov, E., Peeters, R.
  (eds.) Proceedings of the Seventh Workshop on Conformal and Probabilistic
  Prediction and Applications. Proceedings of Machine Learning Research,
  vol.~91, pp. 234--253. PMLR (11--13 Jun 2018),
  \url{http://proceedings.mlr.press/v91/kuleshov18a.html}

\bibitem{liu2008isolation}
Liu, F.T., Ting, K.M., Zhou, Z.H.: Isolation forest. In: Eighth IEEE
  International Conference on Data Mining (ICDM), 2008. pp. 413--422. IEEE
  (2008)

\bibitem{papadimitriou2003loci}
Papadimitriou, S., Kitagawa, H., Gibbons, P.B., Faloutsos, C.: Loci: Fast
  outlier detection using the local correlation integral. In: Proc. of 19th
  International Conference on Data Engineering, 2003. pp. 315--326. IEEE (2003)

\bibitem{pinet2003development}
Pinet, M., Lo, A.: Development of a road weather information system (rwis)
  network for alberta’s national highway system. In: Intelligent
  Transportation Systems (2003)

\bibitem{MF2018}
Rivera, R., Nazarov, I., Burnaev, E.: Towards forecast techniques for business
  analysts of large commercial data sets using matrix factorization methods.
  Journal of Physics: Conference Series  \textbf{1117}(1),  012010 (2018),
  \url{http://stacks.iop.org/1742-6596/1117/i=1/a=012010}

\bibitem{rousseeuw1999fast}
Rousseeuw, P.J., Driessen, K.V.: A fast algorithm for the minimum covariance
  determinant estimator. Technometrics  \textbf{41}(3),  212--223 (1999)

\bibitem{ConformalAD2015}
Safin, A., Burnaev, E.: Conformal kernel expected similarity for anomaly
  detection in time-series data. Advances in Systems Science and Applications
  \textbf{17}(3),  22--33 (2017). \doi{10.25728/assa.2017.17.3.497}

\bibitem{salehi2016smart}
Salehi, M., Zhang, X., Bezdek, J.C., Leckie, C.: Smart sampling: A novel
  unsupervised boosting approach for outlier detection. In: Australasian Joint
  Conference on Artificial Intelligence. pp. 469--481. Springer (2016)

\bibitem{scholkopf2000support}
Sch{\"o}lkopf, B., Williamson, R.C., Smola, A.J., Shawe-Taylor, J., Platt,
  J.C.: Support vector method for novelty detection. In: Advances in neural
  information processing systems. pp. 582--588 (2000)

\bibitem{seni2010ensemble}
Seni, G., Elder, J.F.: Ensemble methods in data mining: improving accuracy
  through combining predictions. Synthesis Lectures on Data Mining and
  Knowledge Discovery  \textbf{2}(1),  1--126 (2010)

\bibitem{Imbalanced2019}
Smolyakov, D., Korotin, A., Erofeev, P., Papanov, A., Burnaev, E.:
  Meta-learning for resampling recommendation systems. In: Eleventh
  International Conference on Machine Vision (ICMV 2018); 110411S (2019). Proc.
  SPIE, vol. 11041 (2019)

\bibitem{OCSVM2018}
Smolyakov, D., Sviridenko, N., Burikov, E., Burnaev, E.: Anomaly pattern
  recognition with privileged information for sensor fault detection. In:
  Pancioni, L., Schwenker, F., Trentin, E. (eds.) Artificial Neural Networks in
  Pattern Recognition. pp. 320--332. Springer International Publishing, Cham
  (2018)

\bibitem{toivonen2001road}
Toivonen, K., Kantonen, J.: Road weather information system in finland.
  Transportation Research Record: Journal of the Transportation Research Board
  \textbf{1741},  21--25 (2001)

\bibitem{ConformalMartingales2017}
Volkhonskiy, D., Burnaev, E., Nouretdinov, I., Gammerman, A., Vovk, V.:
  Inductive conformal martingales for change-point detection. In: Gammerman,
  A., Vovk, V., Luo, Z., Papadopoulos, H. (eds.) Proceedings of the Sixth
  Workshop on Conformal and Probabilistic Prediction and Applications.
  Proceedings of Machine Learning Research, vol.~60, pp. 132--153. PMLR,
  Stockholm, Sweden (13--16 Jun 2017),
  \url{http://proceedings.mlr.press/v60/volkhonskiy17a.html}

\bibitem{zimek2013subsampling}
Zimek, A., Gaudet, M., Campello, R.J., Sander, J.: Subsampling for efficient
  and effective unsupervised outlier detection ensembles. In: Proceedings of
  the 19th ACM SIGKDD international conference on Knowledge discovery and data
  mining. pp. 428--436. ACM (2013)

\end{thebibliography}

\end{document}